\documentclass[10pt,twocolumn,letterpaper]{article}

\usepackage[pagenumbers]{cvpr} 
\usepackage{comment}
\usepackage{bm}
\usepackage[dvipsnames]{xcolor}
\usepackage{graphicx}

\newcommand{\ytodo}[1]{\textcolor{blue}{[#1]}}

\usepackage{indentfirst}
\usepackage{adjustbox}

\definecolor{cvprblue}{rgb}{0.21,0.49,0.74}
\usepackage[pagebackref,breaklinks,colorlinks,citecolor=cvprblue]{hyperref}

\def\paperID{537} 
\def\confName{CVPR}
\def\confYear{2024}

\title{WorDepth: Variational Language Prior for Monocular Depth Estimation}

\author{Ziyao Zeng\textsuperscript{1} \quad
Daniel Wang\textsuperscript{1} \quad
Fengyu Yang\textsuperscript{1} \quad 
Hyoungseob Park\textsuperscript{1} \quad
Yangchao Wu\textsuperscript{2} \\
Stefano Soatto\textsuperscript{2} \quad
Byung-Woo Hong\textsuperscript{3} \quad
Dong Lao\textsuperscript{2} \quad
Alex Wong\textsuperscript{1}
\vspace{3mm} \\
\textsuperscript{1}Yale University \quad \textsuperscript{2}University of California, Los Angeles \quad Chung-Ang University\textsuperscript{3}\\ 
\tt\small \textsuperscript{1}\{ziyao.zeng, 
hyoungseob.park,  
fengyu.yang, 
daniel.wang.dhw33,
alex.wong\}@yale.edu \\
\tt\small  \textsuperscript{2} wuyangchao1997@g.ucla.edu
\tt\small  \textsuperscript{2}\{soatto,lao\}@cs.ucla.edu
\tt\small  \textsuperscript{3}hong@cau.ac.kr
\thanks{Due to an oversight, the author list in the published camera ready does not match that of this manuscript. The above is the correct author list.}
}

\begin{document}

\maketitle

\begin{abstract}
Three-dimensional (3D) reconstruction from a single image is an ill-posed problem with inherent ambiguities, i.e. scale. Predicting a 3D scene from text description(s) is similarly ill-posed, i.e. spatial arrangements of objects described. We investigate the question of whether two inherently ambiguous modalities can be used in conjunction to produce metric-scaled reconstructions. To test this, we focus on monocular depth estimation, the problem of predicting a dense depth map from a single image, but with an additional text caption describing the scene. To this end, we begin by encoding the text caption as a mean and standard deviation; using a variational framework, we learn the distribution of the plausible metric reconstructions of 3D scenes corresponding to the text captions as a prior. To ``select'' a specific reconstruction or depth map, we encode the given image through a conditional sampler that samples from the latent space of the variational text encoder, which is then decoded to the output depth map. Our approach is trained alternatingly between the text and image branches: in one optimization step, we predict the mean and standard deviation from the text description and sample from a standard Gaussian, and in the other, we sample using a (image) conditional sampler. Once trained, we directly predict depth from the encoded text using the conditional sampler. We demonstrate our approach on indoor (NYUv2) and outdoor (KITTI) scenarios, where we show that language can consistently improve performance in both. Code:  \url{https://github.com/Adonis-galaxy/WorDepth}.

\end{abstract}

\section{Introduction}
\label{sec:intro}
The process of imaging is a surjection from a 3D scene to the 2D image domain, where infinitely many 3D scenes can map to the same image. Its inverse problem, estimating the 3D scene structure from a single image, i.e., monocular depth estimation, is therefore ill-posed with inherent ambiguity, such as the scale of the reconstruction. Consequently, induction is necessary, and depth estimation becomes drawing a scene with maximum likelihood from the distribution of all possible scenes, conditioned on the image. This conditional scene distribution is learned by a deep neural network on a chosen training set. While an ideal training set should accurately reflect this distribution, practical challenges arise due to the scarcity of well-established large-scale depth datasets. A crucial question arises: Can any priors, other than the training set, be leveraged to calibrate the learned scene distribution to true real-world statistics? 

\begin{figure}[t]
  \centering
  \vspace{0.3cm}
\includegraphics[width=0.4\textwidth]{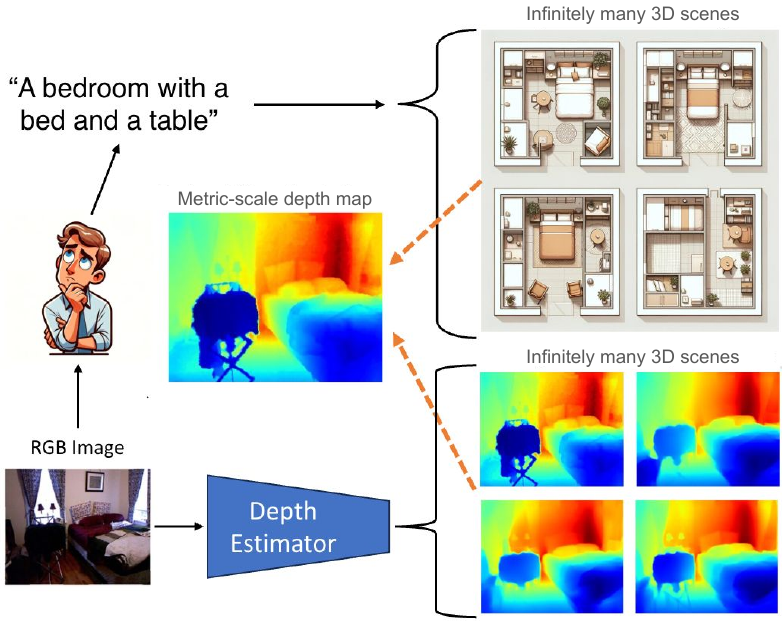}
    \caption{
        \textbf{Language as a prior for depth estimation}. Depth estimation from a single image is an ill-posed problem (i.e., scale), and likewise from text captions (i.e., layout). Can two inherently ambiguous modalities resolve metric-scaled depth estimates?
    }
    \label{fig:teaser}
    \vspace{-0.5cm}
\end{figure}

These priors may come in many forms, from generic priors such as local smoothness and connectivity \cite{garg2016unsupervised,godard2017unsupervised,zhou2017unsupervised,wong2020unsupervised} or object orientation \cite{fei2019geo} that may be imposed as a part of the training objective (regularizer) to specific inductive biases realized as architectural designs (layers) \cite{wong2021unsupervised} or a collection object shapes \cite{fei2018visual}. While generic priors are suitable for a wide variety of scenes, they typically lack specificity, i.e., size or shape of objects within a specific 3D scene. On the other hand, specific network designs may backfire when the assumption motivating the design does not hold, i.e., using specifics about camera parameters for reconstruction. We consider a more flexible source of priors -- language -- that is closely tied to semantics, and often shape (and functionality) \cite{landau1988importance,landau1998object,biederman1988surface}. Consider a text description of ``A bedroom with a bed and a table'' as in Fig.~\ref{fig:teaser}: One can imagine a probable 3D scene containing a bed and a table as the primary objects. In fact, there exist infinitely many 3D scenes compatible with the description, as there are ambiguities in terms of the scene layout and the precise shape of the bed and table. Yet, one may surmise that the scale of the scene is closely related to the objects (and their typical sizes) populating it. This lends to a prior that is specific for a given scene, yet, generic enough without assumptions on the camera used or the shapes within the imaged 3D scene.

Hence, the question at hand becomes whether two inherently ambiguous modalities (camera image and text descriptions) can be exploited for their complementary strengths: In the image, one can observe the layout and object shapes populating the 3D scene; in a text caption, one has strong priors about the scale (and coarse shapes) of the scene. Our work aims to resolve the respective ambiguities of the two modalities by using language to reduce the solution space to yield metric-scaled reconstructions as 2.5D depth maps.

To test the feasibility of this approach, we consider the ill-posed inverse problem of monocular depth estimation, where one predicts a depth map from a single image. Instead of using just an image, we also assume a text description or caption describing the 3D scene captured within the image. Note that we do not make any assumption regarding the source of the description, i.e., it can be dictated by humans or generated by a model. But for practicality, we use an image captioner (ExpansionNet v2 \cite{ExpansionNet_v2}) to generate a brief, concise description of the image.

To exploit the inherent ambiguity of text captions, where a single description can generate infinitely many 3D scenes, we choose to encode the caption using a variational auto-encoder (VAE) as a mean and standard deviation of the plausible scene
layout distribution. By sampling a noise vector from a standard Gaussian and using the reparameterization trick customary in VAEs, we can draw from the latent distribution and decode it into a metric-scaled depth map. Yet, to choose a particular depth map amongst the many possible, one must rely on the image. This is facilitated by a conditional sampler that predicts the noise vector from the given image in place of the one sampled from a Gaussian to be used in the reparameterization step. Consequently, this substitution enables one to sample the most probable depth map, adhering to the scene arrangement and object shapes observed in the image, from the learned distribution. This naturally lends to an alternating optimization process between the (text-)VAE and conditional sampler. 

\begin{figure*}[t!]
  \centering
    \includegraphics[width=1.0\textwidth]{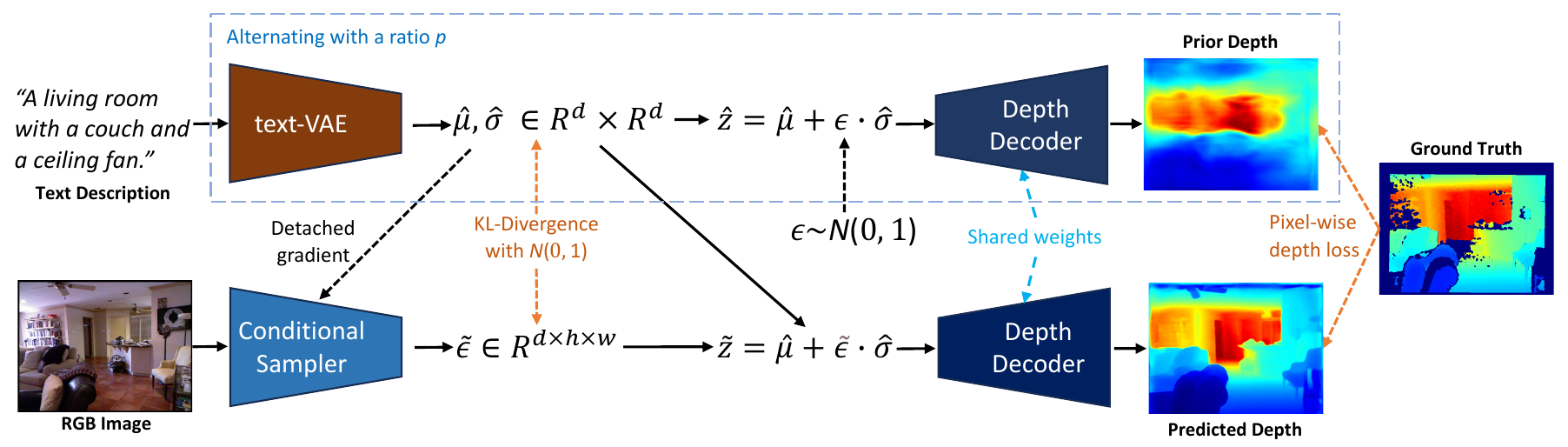}   \caption{\textbf{Training WorDepth.} We begin with optimizing text-VAE by predicting the mean and standard deviation of the latent distribution of depth maps corresponding to a text caption. We then sample $\hat{z}$ from the distribution using the reparameterization trick with $\epsilon \sim \mathcal{N}(0, 1)$ and decode it into a depth map for loss computation. We then optimize a conditional sampler by predicting patch-wise $\tilde{\epsilon}$ from an image to sample $\tilde{z}$ from the latent to yield output depth for the loss computation. The depth decoder is updated in both alternating steps.}
    \label{fig:pipeline}
\end{figure*}

In one alternation, one would predict the mean and standard deviation from the text caption and optimize the text-VAE branch for depth by minimizing a loss with respect to ground truth on the depth map sampled using a standard Gaussian (similar to traditional VAEs). In the other alternation, one would still use the mean and standard deviation predicted by the text-VAE, but instead, use the conditional sampler to ``select'' a specific depth map compatible with the image, and again, minimize a loss on the output depth. Note: that depending on the alternation, either the text-VAE or the conditional sampler is frozen. At test-time, one no longer needs to sample from the Gaussian and may directly predict depth using the text-VAE with the conditional sampler (see \cref{fig:pipeline}). In another mode, one may use the text-VAE alone to generate plausible scenes for a given caption. 

\textbf{Our contributions} are as follows: (i) We propose a variational framework that leverages complementary strengths of two inherently ambiguous modalities for monocular depth estimation; we term our approach, WorDepth. (ii) We introduce an image-based conditional sampler that models the use of language as a conditional prior. (iii) We achieve the state-of-the-art on indoor (NYU Depth V2 \cite{NYU-Depth-V2}) and outdoor (KITTI \cite{KITTI}) benchmarks. (iv) To the best of our knowledge, we are the first to treat language as a variational prior for monocular depth estimation.

\section{Related Work}
\label{sec:rel}

\textbf{Monocular depth estimation } trains by minimizing loss between depth predictions and ground-truth depth maps \cite{chang2021transformer,bhat2021adabins,fu2018deep,qi2018geonet,yin2019enforcing,yang2021transformer,long2021adaptive,lee2019big,ranftl2021vision,upadhyay2023enhancing,yuan2022neural,MonoSwin,wong2020targeted}. 
Specifically, DORN \cite{dorn} employs a spacing-increasing discretization strategy for depth estimation as an ordinal regression problem. 
AdaBins \cite{bhat2021adabins} introduces a transformer block that segments the depth range into adaptive bins.
ASTransformer \cite{chang2021transformer} incorporates an Attention-based Up-sample Block to enhance detailed texture features.  
DepthFormer \cite{li2022depthformer} employs hierarchical aggregation and heterogeneous interaction modules for effective feature affinity and modeling. 
RPSF \cite{mel2022end} presents a differentiable model of the aperture mask.
However, deriving semantics solely from visual cues is challenging because of scale ambiguity and the limited size of fully annotated training datasets. We use language as a prior to ground predictions to metric scale. When ground-truth depth is not available, self-supervised approaches \cite{wu2022toward,ji2021monoindoor,li2021structdepth,yu2020p,bian2021unsupervised,zhao2020towards,zhou2019moving,wang2023sqldepth,wang2023planedepth,peng2021excavating,wong2019bilateral,fei2019geo,zhao2022monovit} rely on geometric constraints, often established via from various modalities, including lidar \cite{yang2019dense,wong2020unsupervised,wong2021adaptive,wong2021learning,liu2022monitored,park2024test,wu2023augundo} and radar \cite{singh2023depth}, or through deliberate design. Arising from training, if done at a large scale, is a prior on the scene that can be exploited for semantic tasks \cite{lao2022depth}. On the other hand, we consider language as a semantic prior to enhance the effectiveness of monocular depth estimation.

\textbf{Variational and generative methods} focus on the ambiguous nature of monocular depth estimation, many involving Diffusion or VAE models for modeling this ambiguity \cite{dikov2022variational,saxena2024surprising,saxena2023monocular,xia2020generating, bloesch2018codeslam, you2024rethinking, liang2023gaufre}.  DepthGen \cite{saxena2023monocular} uses a depth pre-trained diffusion model, which generates depth estimations conditioned on images, and shows that the model is capable of generating multiple plausible depth maps when depth is ambiguous. DDVM \cite{saxena2024surprising} uses a similar approach and designed a training pipeline that can produce both depth maps and optical flow outputs with a diffusion model. \cite{xia2020generating} trained a VAE model that outputs a probability distribution over scene depth given an image, which can then be combined with additional inputs for more accurate depth estimations. VDN \cite{dikov2022variational} models depth as a distribution with its variance interpreted as uncertainty. The CodeSLAM model \cite{bloesch2018codeslam} also employed a VAE conditioned on image intensities for depth estimation. However, although these work explored the idea of uncertainty in depth estimation, and even combined other modalities of inputs \cite{xia2020generating}, none have experimented with language priors, and most VAE-based approaches use images to obtain the mean of the modeled distribution, which is fundamentally different from WorDepth.

\textbf{Foundation models}~\cite{CLIP,Blip,Blip-2,Dino,Dinov2, girdhar2023imagebind, zhu2023languagebind, guo2023point, yang2024unitouch, pan2024efficient, zheng2024iterated} acquire a comprehensive understanding of languages, images, and other data types through pre-training under substantial and diverse datasets, thus forming an effective baseline for downstream tasks~\cite{yang2022touch, yang2023generating, li2023isolated, dou2024tactile, DSPoint, yang2022sparse, iQuery, you2022mine, Wu_2023_boosting,bhat2021adabins, you2023implicit, Liang2023SAFF, Zhao2022RBCRT, zhang2023llama, zhang2024mathverse}. To leverage foundation models for monocular depth estimation, TADP \cite{TADP} uses captions created by AI to enhance the correlation between text and images in diffusion-based vision models. VPD \cite{VPD} leverages a diffusion-based pipeline with cross-attention between text and images.  Dinov2 \cite{Dinov2} trains a ViT \cite{ViT} with 1B parameters using an automatically built image dataset under contrastive learning objectives. 
Unlike methods that rely on foundation models for feature extraction, WorDepth is potentially more efficient for industrial applications.

\textbf{Vision-language models} are designed to build connections between visual and language inputs. 
CLIP \cite{CLIP} conducts contrastive learning between text-image pairs, empowering various tasks like few-shot image classification ~\cite{gao2021clip_cls,zhang2021vt_cls,zhang2021tip_cls,zhou2021learning_cls}, image segmentation ~\cite{zhou2021denseclip_seg,rao2021denseclip_detect_seg}, object detection ~\cite{zhou2022detecting_detect,rao2021denseclip_detect_seg}, and 3D perception \cite{PointCLIP,PointCLIP_v2,DepthCLIP,DepthCLIPv2}. In light of the powerful emerging ability brought by recent vision-language models, some works have tried to apply the vision-language model for monocular depth estimation. 
DepthCLIP~\cite{DepthCLIP} leverages the semantic depth response of CLIP~\cite{CLIP} with a depth projection scheme to conduct zero-shot adaptation from the semantic language response to monocular depth estimation. 
Furthermore, \cite{DepthCLIPv2} extends DepthCLIP with learnable prompts and depth codebook to narrow the depth domain gap among different scenes. 
Likewise, \cite{DepthCLIP_auty} modifies DepthCLIP~\cite{DepthCLIP} using continuous learnable tokens in place of discrete human-language words.
Additionally, VPD \cite{VPD} exploits the high-fidelity embedding of a pre-trained text-to-image diffusion model in monocular depth estimation.
However, existing methods using vision-language models rely on implicit modeling. Conversely, WorDepth explicitly models language as a prior for depth estimation and exploits strong priors regarding the size of objects described in text captions to better ground monocular depth (often scaleless) to metric scale.

\section{Method}
\label{sec:method}

Given an RGB image $x : \Omega \subset \mathbb{R}^2 \rightarrow \mathbb{R}^3$, monocular depth estimation aims to infer a dense depth map $y : \Omega \subset \mathbb{R}^2 \rightarrow \mathbb{R}_+$ using a parameterized function $h$ realized as a neural network, i.e., $y := h(\cdot)$. We consider a supervised dataset $\mathcal{D} = \{x^{(m)}, \textbf{t}^{(m)}, y^{*(m)}\}_{m=1}^M$ with $M$ samples, where $y^* : \Omega \subset \mathbb{R}^2 \rightarrow \mathbb{R}_+$ denotes the ground-truth depth map, and $\textbf{t}$ the text caption describing the image. 

\subsection{Text variational auto-encoder}
\label{sec:text-vae}
To incorporate language priors to monocular depth estimation, we first design a variational auto-encoder (VAE) to learn the latent distribution of possible depth maps as described by the text caption. This VAE is comprised of the text encoder from a pre-trained vision-language model, CLIP \cite{CLIP}, which by default offers a shared latent space between vision and text embeddings, followed by a multi-layer perceptron (MLP) to estimate the mean $\hat{\mu} \in \mathbb{R}^d$ and standard deviation $\hat{\sigma} \in \mathbb{R}^d$ of the latent distribution of plausible scenes based on the text encoding. Note that the CLIP text encoder is frozen at all times and never updated when training WorDepth. Specifically, given a text caption $\textbf{t} = \{ t_1, t_2, ... \}$, we first encode it using the CLIP text encoder and estimate the mean and standard deviation as $(\hat{\mu}, \hat{\sigma}) = g_\psi(\textbf{t}) \in \mathbb{R}^{2 \times d}$ using a multi-layer perceptron (MLP). To sample from the distribution parameterized by $\hat{\mu}$ and $\hat{\sigma}$, we first draw a noise vector $\epsilon \in \mathbb{R}^d$ from a standard Gaussian $\epsilon \sim \mathcal{N}(0, 1)$. Then, we use $\epsilon$ to sample from the latent distribution via the reparameterization trick \cite{kingma2013auto}, $\hat{z} = \hat{\mu} + \epsilon \cdot \hat{\sigma}$. We refer to this module as a text variational auto-encoder (text-VAE). To generate a depth map $\hat{y}$ from the sample $\hat{z} \in \mathbb{R}^d$, we first duplicate $\hat{z}$ along the horizontal and vertical axes to yield a $d \times h \times w$ latent (choice of design to be discussed below in \cref{sec:conditional-sampler}) and feed it through a depth decoder to yield $\hat{y} = h_\phi(\hat{z}) \in \mathbb{R}_+^{H \times W}$, where we overload $\hat{z}$ as the spatially duplicated latent, and $H$ and $W$ denote the height and width of the depth map, preset as hyperparameters to match the desired image dimensions.  

To train our text-VAE and depth decoder, we minimize
\begin{equation}
\begin{aligned}
    \mathcal{L}_\text{VAE} = \mathcal{L}_\text{SI}(y^{*}, \hat{y}) + \alpha \cdot \mathcal{L}_\text{KL}(\hat{\mu}, \hat{\sigma})
\end{aligned}
\label{eq:loss_vae}
\end{equation}
with respect to $\psi$ and $\phi$, where $\mathcal{L}_\text{SI}$ is the scale invariant loss (\cref{eqn:si-loss}), $\mathcal{L}_\text{KL}$ the KL divergence loss (\cref{eqn:kld-loss}) as detailed in Section~\ref{sec:loss}, and $\alpha$ the weight of the KL divergence term.

\subsection{Image-based conditional sampler}
\label{sec:conditional-sampler}

While our text-VAE can predict plausible metric-scaled depth maps from text captions, we are interested in the depth map corresponding to a specific image. To do so, we treat text-VAE as the latent prior distribution of the plausible scene layouts. Predicting depth $\tilde{y}$ for a specific image $x$ requires \emph{sampling} the latent corresponding to the depth map of the 3D scene layout with the highest likelihood to be compatible with the observed image, i.e., prior conditioned on the image. To this end, we introduce an image-based conditional sampler that will predict the sample $\tilde{\epsilon}$ in place of $\epsilon \sim \mathcal{N}(0, 1)$ drawn from the standard Gaussian. Using the reparameterization trick as before, we will use $\tilde{\epsilon}$ to select the latent vector $\tilde{z}$ to be decoded by the depth decoder.

Specifically, our image-based conditional sampler utilizes a Swin-L transformer backbone to encode an image $x \in \mathbb{R}^{3 \times H \times W}$. We chose this design to exploit the locality of the tokens produced by Swin-L. The tokens are then encoded into $h \times w$ number of local samples $\tilde{\epsilon} \in \mathbb{R}^{d \times h \times w}$ to be used to sample from the latent distribution of our text-VAE; in other words, we perform ``patch-wise'' selection from latent distribution for more granular predictions. To do so, we additionally include $\hat{\mu}$ and $\hat{\sigma}$ as part of its input. We note that $\hat{\mu}$ and $\hat{\sigma}$ have been detached from the computational graph and treated as input. We refer to this module as our conditional sampler $\tilde{\epsilon} = f_{\varphi}(x, \hat{\mu}, \hat{\sigma})$, which aims to estimate the most probable latent variable of text-VAE. Thus, the scene layout latent vector is now given by $\tilde{z} = \hat{\mu} + \tilde{\epsilon} \cdot \hat{\sigma}$, and the predicted depth $\tilde{y} = h_\phi(\tilde{z})$. As an implementation detail, we note that skip connections from the encoder $f_\varphi$ are injected into $h_\phi$ by concatenation; when training text-VAE (\cref{sec:text-vae}), feature maps of skip connections are of the same size, but populated with zeros instead.

To train the conditional sampler, we minimize the same loss (\cref{eq:loss_vae}) as that of text-VAE:
\begin{equation}
\begin{aligned}
    \mathcal{L}_\text{CS} = \mathcal{L}_\text{SI}(y^{*}, \tilde{y}) + \beta \cdot \mathcal{L}_\text{KL}(\tilde{\mu}, \tilde{\sigma})
\end{aligned}
\end{equation}
with respect to $\varphi$ and $\phi$. With a batch size of $b$, the number of $\tilde{\epsilon}$ is $b \times h \times w$, while $\tilde{\mu}$ and $\tilde{\sigma}$ are the sample mean and standard deviation of $\tilde{\epsilon}$ over a batch.  We impose a KL divergence loss as regularization so that the estimated $\tilde{\epsilon}$ does not drift from the standard Gaussian, which also serves to improve training stability.

\subsection{Training Loss}
\label{sec:loss}
\begin{figure*}[t!]
  \centering
    \includegraphics[width=\textwidth]{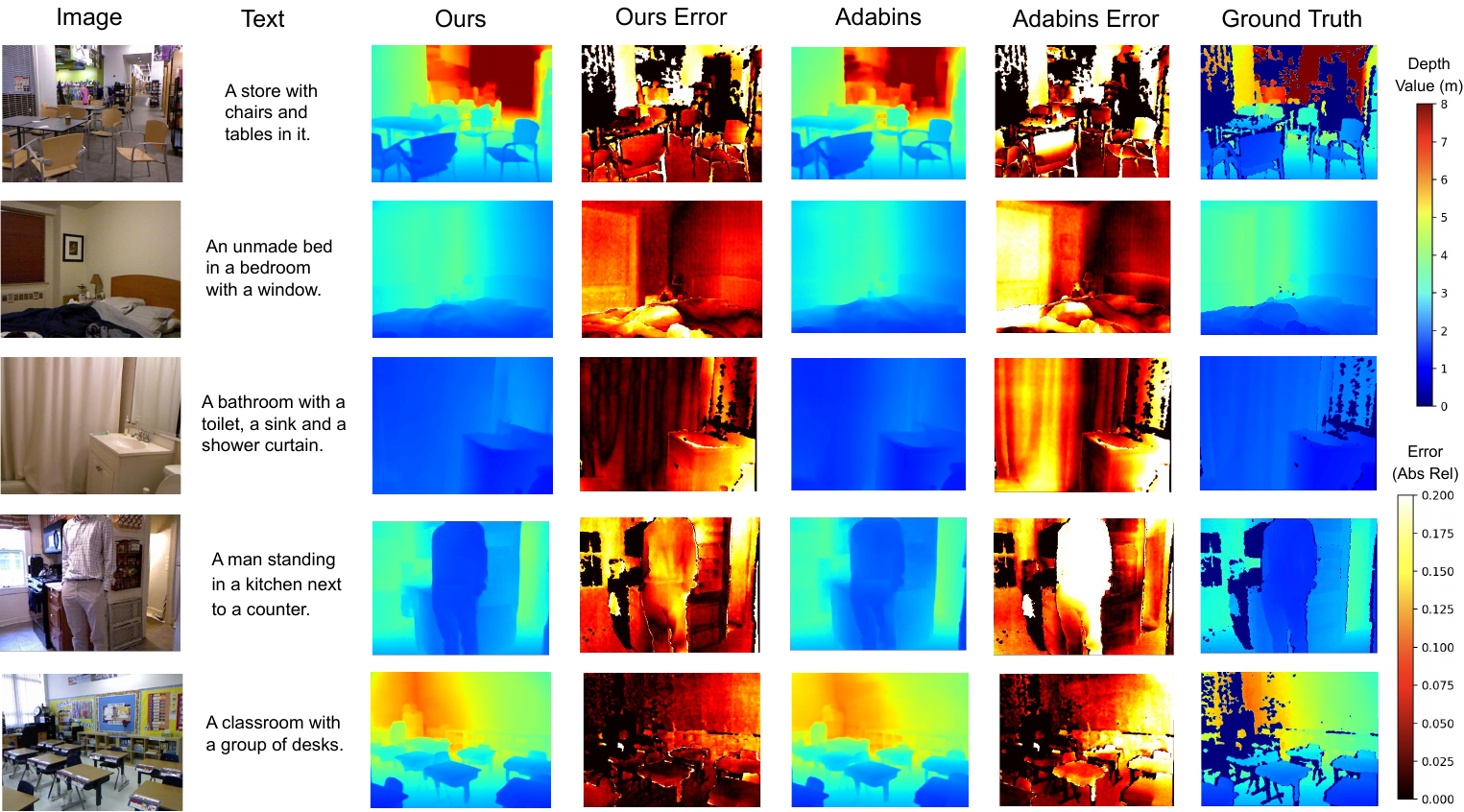}   \caption{\textbf{Qualitative results on NYU Depth V2.} We compare WorDepth with AdaBins \cite{bhat2021adabins}. Text descriptions are generated using ExpansionNet v2 \cite{ExpansionNet_v2}. Overall, WorDepth improves uniformly across the image (darker in error map), implying better scale. WorDepth also predicts more accurate depth in regions corresponding to ``chairs'', ``window'', ``shower curtain'', ``man'', and ``desks'', which are objects specified by text descriptions. Note: Zeros in the ground truth depth map indicate the absence of valid depth values.}
    \label{fig:vis_nyu}
    \vspace{-2mm}
\end{figure*}

\textbf{Scale invariant loss.} We minimize a supervised loss using ground truth $y^*$. To improve training stability over diverse scenes, we use the scale-invariant depth loss \cite{Eigen-Split}:
\begin{equation}
\label{eqn:si-loss}
\mathcal{L}_\text{SI} (y, y^*) =\frac{1}{N_{e}} \sum_{(i,j) \in \Omega} e(i, j)^2-\frac{\gamma}{N_{e}^2}(\sum_{(i,j) \in \Omega} e(i, j))^2,
\end{equation}
where $e(i, j)=\log y(i,j)-\log y^*(i,j)$, $\Omega$ denotes the image space, $N_e$ the number of pixels, $y$ the predicted depth, and $\gamma$ the scaling factor to control the sensitivity of the loss.

\textbf{Kullback-Leibler (KL) divergence loss.} Following \cite{kingma2013auto}, we employ the KL Divergence loss as a regularizer, which biases the predicted latent distribution (parameterized by mean $\mu$ and standard deviation $\sigma$) towards a standard Gaussian distribution. We apply the Kullback-Leibler divergence loss to $\mu$ and $\sigma$ as follows:
\begin{equation}
\label{eqn:kld-loss}
\mathcal{L}_\text{KL}(\mu, \sigma) = -\log(\sigma) + \frac{\sigma^2 + \mu^2}{2} - \frac{1}{2}.
\end{equation}

\subsection{Optimizing Wordepth}
\label{sec:optimization}
Training Wordepth involves optimizing text-VAE with our conditional sampler: One may choose to first train text-VAE until convergence (i.e., optimize for $\psi^*, \phi^*$), then freeze $\psi^*, \phi^*$, and finally train the image-based conditional sample (i.e., optimize for $\varphi^*$). However, we find that doing so often results in the conditional sampler being trapped in a suboptimal local minimum. Moreover, this introduces the inconvenience of an extra stage of training. Instead, we propose an alternating optimization scheme to train the text-VAE with conditional sampler. In one alternating step, we freeze the conditional sampler and train the text-VAE and depth decoder following the procedure in \cref{sec:text-vae}, i.e., predicting $\hat{\mu}$ and $\hat{\sigma}$ from text caption $\textbf{t}$ and using the reparameterization trick with an $\epsilon$ drawn from a standard Gaussian to sample the latent vector. In the next alternating step, we freeze text-VAE and train the conditional sampler with the depth decoder following \cref{sec:conditional-sampler}, i.e., predicting $\hat{\mu}$ and $\hat{\sigma}$ using the frozen text-VAE and sample from the latent distribution using $\tilde{\epsilon}$ predicted from the image. These alternating steps are repeated with a ratio of $p$ (for optimizing text-VAE) to $1-p$ (for optimizing the conditional sampler). 

\textbf{Inference.} Once trained, we no longer require drawing $\epsilon$ from a standard Gaussian. Instead, at test time, the inference step simply follows \cref{sec:conditional-sampler}. In another mode, if one wants to generate depth maps from text captions, one can discard the conditional sampler branch and directly sample from a standard Gaussian instead.

\section{Experiments}
\label{sec:experiments}

\textbf{Datasets.}
We evaluate our method on indoor (NYU Depth V2~\cite{NYU-Depth-V2}) and outdoor (KITTI~\cite{KITTI}) scenarios. NYU Depth V2 contains 480$\times$640 images  with depth values from $1 \times 10^{-3}$ to 10 meters. We follow \cite{big-to-small} for the dataset partition, which contains 24,231 train images and 654 test images. KITTI contains 352$\times$1216 images where depth values from $1 \times 10^{-3}$ to 80 meters. 
We adopt the Eigen Split~\cite{Eigen-Split} consisting of 23,488 training images and 697 testing images. Following~\cite{yuan2022neural,bhat2021adabins}, after cleaning out samples without valid ground truth, we have 652 valid images for testing.

\begin{table*}[t!]
\centering
\footnotesize
\begin{adjustbox}{width=0.95\linewidth}
\begin{tabular}{l|c|ccc|ccc}
\toprule
Method & Backbone & \textbf{$\delta<1.25 \uparrow$} & \textbf{$\delta<1.25^{2} \uparrow$} & $\delta<1.25^{3} \uparrow$ & Abs Rel $\downarrow$ & $\log _{10} \downarrow$ & RMSE $\downarrow$ \\ \midrule
DepthCLIP \cite{DepthCLIP} &CLIP (zero-shot) &0.394   &0.683   &0.851	&0.388  &0.156   &1.167 \\

CLIPMDE \cite{DepthCLIP_auty} &CLIP &0.465 & 0.776 & 0.922 &
0.319 & 0.139  &  0.970 \\

GeoNet \cite{qi2018geonet} & ResNet-50  & 0.834 & 0.960 &0.990              & 0.128 &0.057       & 0.569\\

DORN \cite{dorn}            &ResNet-101  & 0.828 & 0.965 & 0.992        & 0.115 & 0.051 & 0.509\\

\citet{yin2019enforcing} & ResNeXt-101& 0.875 & 0.976 &0.994             &0.108  &0.048    &0.416\\

TransDepth \cite{yang2021transformer} & ViT-B  & 0.900 & 0.983         &0.996  &0.106 & 0.045  &0.365\\

ASN \cite{long2021adaptive} & HRNet-48 & 0.890 & 0.982 & 0.996              &0.101 & 0.044   &0.377\\

Big to Small \cite{lee2019big} & DenseNet-161 & 0.885 & 0.978 &0.994                 &0.110 & 0.047    &0.392\\

DPT-Hybird \cite{ranftl2021vision} & ViT-B & 0.904 & 0.988 &\textbf{0.998}           &0.110  &0.045    &0.357\\

ASTransformer \cite{chang2021transformer} &ViT-B &0.902 &0.985 &0.997  &0.103 &0.044 &0.374\\

AdaBins \cite{bhat2021adabins} & EffNet-B5 + ViT-mini & 0.903 & 0.984 &0.997   &0.103 & 0.044   &0.364\\

NeWCRFs \cite{yuan2022neural} & Swin-L & 0.922 & \textbf{0.992} &\textbf{0.998}   &0.095   &0.041  &0.331\\

\citet{MonoSwin} &Swin-L &0.921 &0.990 &\textbf{0.998}   &0.093 &0.040 &0.331 \\

DepthFormer \cite{li2022depthformer}  &Swin-L &0.923 	&0.989	&0.997      &0.094	&0.040 &0.329\\

\midrule

Baseline        & Swin-L &0.910   &0.990   &\textbf{0.998}	&0.098  &0.043   &0.351\\

\textbf{WorDepth} & Swin-L & \textbf{0.932}  & \textbf{0.992} & \textbf{0.998} & \textbf{0.088} & \textbf{0.038} & \textbf{0.317} \\

\textbf{\%Improvement} & - &  {+2.42\%} & {+0.02\%} & {+0.00\%} & {-10.20\%} & {-11.63\%} & {-9.69\%}\\
   
\bottomrule
\end{tabular}
\end{adjustbox}
\caption{\textbf{Quantitative results on NYU Depth V2.} The baseline method is to directly train a Swin-L image encoder and the depth decoder without the help of language prior. Improvement refers to the performance enhancement relative to the Baseline.}
\label{tab:nyu_results}
\vspace{-3mm}
\end{table*}

\textbf{Network Architecture.} We use the ResNet-50 \cite{resnet} version of CLIP \cite{CLIP} text encoder to extract text features. We use ExpansionNet-v2 \cite{ExpansionNet_v2} for captioning images for efficiency. We set the dimension $d$ of the latent space of the text-VAE and image-based conditional sampler to be 128. As for the image-based conditional sampler, we use a Swin-L Transformer backbone \cite{liu2021swin} pre-trained on ImageNet \cite{deng2009imagenet}. For the text-VAE, given CLIP features of size 1024, we use a 3-layer MLP with hidden dimensions of 512, 256, and 128 to encode text features. For the depth decoder, there are 3 convolutional up-sampling and refinement layers. For depth prediction, we attach 3 skip connections from the conditional sampler to the depth decoder between corresponding layers. 
When optimizing for text-VAE by our alternating optimization scheme (\cref{sec:optimization}), we sample $\epsilon \sim \mathcal{N}(0, 1)$ from a standard Gaussian; as an implementation detail, all values passed from the skip connections are set to be zero.

\textbf{Hyperparameters.} We use the Adam \cite{kingma2014adam} optimizer without weight decay. The learning rate is reduced from $3 \times 10^{-5}$ to $1 \times 10^{-5}$ by a cosine learning rate scheduler. The model is trained for 50 epochs for both KITTI and NYU Depth V2 under this scheduler. $\gamma$ for scale-invariant loss is set to 0.85, and the weights $\alpha$ and $\beta$ for KL-Divergence are set to $1 \times 10^{-3}$. We set the probability $p$ to optimizing text-VAE branch to 1\%. Data augmentation includes random gamma within $[0.9, 1.1]$, random brightness within $[0.75, 1.25]$ for NYU Depth V2~\cite{NYU-Depth-V2} and $[0.9, 1.1]$ for KITTI~\cite{KITTI}, random color intensity within $[0.9, 1.1]$ for each channel, random horizontal flipping with 50\% probability, and random rotations within $[-2.5, 2.5]$ degrees.

\begin{figure*}[t!]
  \centering
    \includegraphics[width=\textwidth]{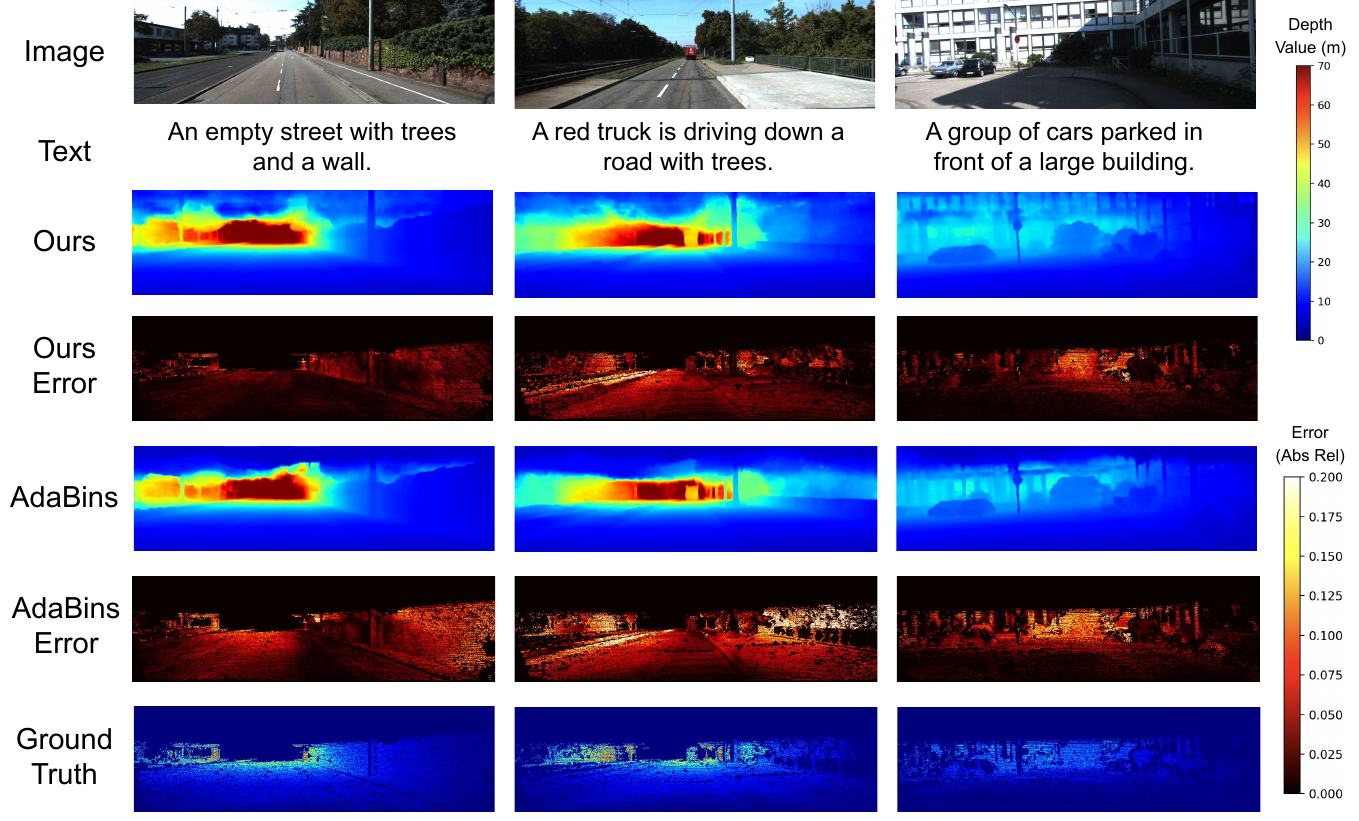}
    \vspace{-6mm}
        \caption{\textbf{Visualization of depth estimations on KITTI.} Compared with AdaBins \cite{bhat2021adabins}, WorDepth improves uniformly across the image (darker in error map), implying better scale. WorDepth also predicts more accurate depth in regions corresponding to ``wall'', ``trees'', ``building'', which are objects specified by text descriptions. Note: Zeros in ground truth depth indicate the absence of valid depth values.}
    \label{fig:vis_kitti}
    \vspace{-2mm}
\end{figure*}

\textbf{Evaluation metrics.}
Following \cite{Va-depthnet,chang2021transformer}, we evaluate WorDepth and baseline methods quantitatively using mean absolute relative error (Abs Rel), root mean square error (RMSE), absolute error in log space $\left(\log _{10}\right)$, logarithmic root mean square error ($\text{RMSE}_{\log}$) and threshold accuracy $\left(\delta_{i}\right)$. The evaluation metrics are summarized in the Supp. Mat. For qualitative results and comparisons, see  \cref{fig:vis_nyu} and \ref{fig:vis_kitti}, where the error map shows the absolute relative error.

\textbf{Quantitative results.}
We show results on NYU Depth V2 in \cref{tab:nyu_results}, where we improve over the baseline and existing works across all evaluation metrics. We want to highlight that WorDepth significantly excels in terms of the threshold accuracy $\delta<1.25$, which measures the proportion of predictions deviating from the ground truth within a specific range. 
We note that while existing methods often produce high fidelity shapes (i.e., ordinal relationships of points) in depth maps, the scale tends to be off -- leading to lower $\delta<1.25$. Our gain in the $\delta<1.25$ metric indicates that a greater proportion of depth estimations align closely with the ground truth, thanks to better scale estimated based on objects that populate the scene, thereby yielding depth values in ranges closer to that of ground truth. 

\cref{tab:kitti_eigen_results} shows the results on the KITTI dataset, using the Eigen Split \cite{Eigen-Split} partition. WorDepth also achieves state-of-the-art performance.  Like NYU Depth V2, WorDepth improves the threshold accuracy $\delta<1.25$, however, the relative performance gain on this metric is not as pronounced as on NYU Depth V2. 
This difference can be due to the wider range of object sizes and shapes that may populate an outdoor scene that are attributed to the same equivalence class of an object. For example, the term ``car'' may refer to a sedan, a coupe, or a hatchback -- all exhibit different sizes (coupes are smaller than sedans) and shapes (hatchbacks have an elevated and connect trunk). While text captions give flexibility between generality and specificity as a prior, in cases where captions tend to be vague, the explicit reliance (by modeling as a conditional prior) on them may backfire, leading to incorrect shapes and sizes. Nonetheless, conditioning on the image resolve such cases to a degree and usage of the prior leads to more benefits than harm.

\textbf{Qualitative comparisons.}
\label{sec:vis}
We show representative visual examples comparing WorDepth with a baseline method AdaBins \cite{bhat2021adabins} on the NYU Depth V2 dataset in \cref{fig:vis_nyu}, to highlight the benefit of the language prior.

From the error map where brighter regions indicate larger errors, it is evident that WorDepth predicts more accurate depth for objects mentioned in the text description, like ``chairs and tables'' in the first row, ``a window'' in the second row, ``a shower curtain'' in the third row, ``a man'' in the fourth row, and ``a group of desks'' in the last row. 
Note that errors maps of WorDepth shows improvement uniformly across the image regions; this implies that our method estimates a better scale than existing ones, thanks to priors about object size (and coarse shapes) from text captions. Knowing that a certain object exists within an image reduces the problem to ``placing'' the object in the 3D scene based on its shape and location in the image. We showed that scale can be inferred from language, which can narrow down the solution space of depth prediction, leading to improved accuracy. 

A similar pattern is also evident in KITTI (\cref{fig:vis_kitti}). 
Examples include improved accuracy for ``a wall" shown in the first column, ``trees" in the second column, and ``a group of cars" alongside ``a large building" in the last column. 
This observation is intriguing because, for example, the text ``a wall" is ambiguous by itself, especially in outdoor scenes, where the wall could be any size or distance away from the camera, 1 or 100 meters. However, the text description of a scene, either from a human annotator or a deep neural network, inherently carries biases that emphasize ``a wall" when its size (tall or wide enough) or depth falls within a specific range while ignoring it when it falls within another range. The resulting prior embedded in the text description may convey more scale information than initially apparent.

\begin{table*}[t!]
\centering
\footnotesize
\begin{adjustbox}{width=0.95\linewidth}
\begin{tabular}{l|c|ccc|ccc}
\toprule
Method & Backbone & $\delta<1.25 \uparrow$ & $\delta<1.25^{2} \uparrow$ & $\delta<1.25^{3} \uparrow$ & Abs Rel $\downarrow$ & $\text{RMSE}_{\log} \downarrow$ & RMSE $\downarrow$ \\ \midrule
CLIPMDE \cite{DepthCLIP_auty} &CLIP &0.550 &0.830 &0.938 &0.303 &0.119 &6.322 \\
DORN \cite{dorn}            &ResNet-101  &0.932 &0.984 &0.995  &0.072  &0.120 &2.727\\

\citet{yin2019enforcing} & ResNeXt-101  &0.938 &0.990 &0.998 &0.072 &0.117 &3.258 \\

TransDepth \cite{yang2021transformer} & ViT-B  &0.956 &0.994 &\textbf{0.999} &0.064 &0.098 &2.755 \\

Big to Small \cite{lee2019big} & DenseNet-161 &0.955 &0.993 &0.998 &0.060 &0.096 &2.798\\

DPT-Hybird \cite{ranftl2021vision} & ViT-B &0.959 &0.995 &\textbf{0.999} &0.062 &0.092   &2.573\\

ASTransformer \cite{chang2021transformer} &ViT-B  &0.963 &0.995 &\textbf{0.999} &0.058 &0.089 &2.685\\

AdaBins \cite{bhat2021adabins} & EffNet-B5+ViT-mini & 0.964 &0.995 &\textbf{0.999}   &0.058 &0.089  &2.360\\

NeWCRFs \cite{yuan2022neural} & Swin-L & 0.974 &0.997 &\textbf{0.999}   &0.052   &0.079  &2.129\\

\citet{MonoSwin} &Swin-L &0.972 &0.996 &\textbf{0.999} &0.054 &0.081  &2.134  \\

DepthFormer \cite{li2022depthformer}  &Swin-L &0.975    &0.997  &\textbf{0.999}  &0.052  &0.079  &2.143 \\

\midrule

Baseline        & Swin-L &0.969  &0.996   &\textbf{0.999}   &0.054   &0.085  &2.343\\

\textbf{WorDepth} & Swin-L & \textbf{0.979} & \textbf{0.998} & \textbf{0.999}   & \textbf{0.049}  & \textbf{0.074}  & \textbf{2.039} \\
\textbf{\% Improvement} & - &  {+1.03\%} & {+0.20\%} & {+0.00\%} & {-9.26\%} & {-12.94\%} & {-12.97\%}\\
   
	\bottomrule
	\end{tabular}
\end{adjustbox}
\caption{\textbf{Quantitative results on KITTI Eigen Split.} The baseline method is to directly train a Swin-L image encoder and the depth decoder without the help of language prior. Improvement is the relative performance gain compared with the Baseline.}
\vspace*{-4mm}
\label{tab:kitti_eigen_results}
\end{table*}

\begin{table}[t]
\centering
\begin{adjustbox}{width=\linewidth}
\LARGE
\setlength{\tabcolsep}{2pt}
{
	\begin{tabular}{l|ccc|ccc}
	\toprule

		$p$& $\delta<1.25\uparrow$& $\delta<1.25^{2}\uparrow$& $\delta<1.25^{3}\uparrow$& AbsRel$\downarrow$ & $\log _{10}\downarrow$ & RMSE$\downarrow$ \\ 
  
        \midrule
        0\%    &0.929 &0.990 &\textbf{0.998} &0.091 &0.039 &0.323\\

        \textbf{1\%}   &\textbf{0.932}  & \textbf{0.992} & \textbf{0.998} & \textbf{0.088} & \textbf{0.038} & \textbf{0.317}\\

        10\%  &0.930  &0.991  &\textbf{0.998} &0.090  &0.039  &0.322              \\

        50\%  &0.763  &0.942  &0.987 &0.163  &0.068  &0.527               \\
  
        90\%  &0.642  &0.906  &0.975 &0.211  &0.089  &0.687             \\

        100\%  &0.590  &0.889  &0.973 &0.225  &0.097  &0.746             \\


	\bottomrule
	\specialrule{0em}{7pt}{7pt}
	\end{tabular}
 }
\end{adjustbox}
\vspace*{-3mm}
\caption{\textbf{Sensitivity to different ratios of alternating optimization steps between text-VAE and conditional sampler on NYU Depth V2.} $p$ denotes probability of optimizing text-VAE. While more steps spent on text-VAE will yield better generative results, it comes at the cost of slower convergence for the sampler.}
\label{tab:abaltion}
\vspace*{-3mm}
\end{table}

\textbf{Optimizing with different alternation ratios.}
\label{sec:ablation}
As a sensitivity study, we investigate how different ratios of alternating optimization steps between text-VAE and conditional sampler have an effect on the performance of WorDepth.
We find that optimizing text-VAE with a lower ratio will lead to a more deterministic model, which is anticipated. On the other hand, optimizing text-VAE more frequently enables the model to learn a better variational prior on the depth maps corresponding to text captions, which facilitates the generation of diverse prior depth maps. However, this comes at the cost of training time as the conditional sampler takes more steps to converge and, given a fixed number of steps, results in more blurry predictions. We identify the ratio at 1\% in updating text-VAE to be the best empirically (\cref{tab:abaltion}). Ratios exceeding 10\% notably degrades performance given a fixed training length because of fewer updates to the sampler. Note that at 100\%, where we do not condition the image, caption to depth generation still yields reasonable results as the text captions produce plausible statistics that match the ground truth depth. On the other hand, without the modeling language as a variational prior (at 0\%, where we train both text-VAE and conditional optimizer jointly as a direct map from single image and caption to depth), performance degrade to do the lack of the prior.

\textbf{Zero-shot Generalization.}
\label{sec:zero_shot}
Given the smaller domain gap in language descriptions across different scenes compared to images, we conduct a zero-shot transfer experiment to highlight our improved generalization ability. 
We train the model on the NYU Depth V2~\cite{NYU-Depth-V2} and test it on the Sun-RGBD~\cite{Sun-RGBD} without fine-tuning. As shown in \cref{tab:zero_shot}, WorDepth outperforms baseline methods by a substantial margin, indicating the transferability of language priors which underscores the robustness of text descriptions in handling scene variability. This suggests that language descriptions may offer a more stable basis for generalization across diverse data domains than direct visual signals.

\section{Discussion}
\label{sec:conclusion}
In this study, we seek to answer the question of whether language can be used to calibrate the learned scene distribution to true real-world statistics. The answer is yes, which is valuable for 
circumventing the long-standing problem of scale ambiguity in monocular depth or structure-from-motion problems. The approach is a first in leveraging complementary properties of two modalities with inherent ambiguities for the 3D reconstruction, to address the deficits in one another. We show that by exploiting the layout/scene ambiguity in language as a strength via our variational approach, we can ground predictions to metric scale. This opens up new avenue in how one can address the issue of scale in 3D reconstruction as well as provide a direct framework to extending the many works that currently are limited to relative or scaleless depth predictions. 

\begin{table}[t]
\hspace*{-3mm}
\centering
\begin{adjustbox}{width=1.05\linewidth}
{\LARGE
\setlength{\tabcolsep}{1pt}
\begin{tabular}{l|ccc|ccc}
\toprule
Method& \textbf{$\delta<1.25\uparrow$} & \textbf{$\delta<1.25^{2}\uparrow$} & $\delta<1.25^{3}\uparrow$ & AbsRel$\downarrow$ & $\log _{10}\downarrow$ & RMSE$\downarrow$ \\ \midrule

Adabins &0.771   &0.944   &0.983	&0.159  &0.068   &0.476\\

DepthFormer & 0.815 & 0.970 & 0.993 & 0.137  & 0.059 &0.408\\
\hline
Baseline     &0.803   &0.965   &0.990	&0.141  &0.062   &0.427\\

\textbf{WorDepth}   & \textbf{0.833}  & \textbf{0.976} & \textbf{0.994} & \textbf{0.123} & \textbf{0.054} & \textbf{0.376} \\

\bottomrule
\end{tabular}
}
\end{adjustbox}

\caption{\textbf{Zero-shot generalization to SUN-RGBD.} The models are trained on the NYU Depth V2 and testing on the Sun-RGBD without any fine-tuning.}
\vspace{-4mm}
\label{tab:zero_shot}
\end{table}

\textbf{Limitations.} Generic regularizers typically yield little gains, but do little harm; specific regularizers can provide larger boosts but are limited in their applications.  While using language as a prior gives flexibility between the two, specificity in the caption controls the degree of regularization imposed. Naturally, vague captions give little to no information on object shape or size, so there is little to be gained; specific, but incorrect captions may misfire, barring any malicious intent. As WorDepth relies on the quality of the caption, it is susceptible to inaccuracies stemming from descriptions provided by the image captioner. Its ease of use also opens up vulnerabilities from malicious users who may choose captions to steer predictions incorrectly.

\noindent\textbf{Acknowledgements.} This work was supported by NSF 2112562 Athena AI Institute.

{
    \small
    \bibliographystyle{ieeenat_fullname}
    \bibliography{main}
}

\clearpage
\setcounter{page}{1}
\maketitlesupplementary
\appendix

\section{Evaluation metrics.}
\label{sec:eval_metrics}
Drawing on \cite{Va-depthnet,chang2021transformer}, our evaluation of WorDepth alongside comparison methods involves a quantitative assessment through several metrics. These include mean absolute relative error (Abs Rel), root mean square error (RMSE), absolute error in log space $\left(\log _{10}\right)$, logarithmic root mean square error ($\text{RMSE}_{\log}$) and threshold accuracy $\left(\delta_{i}\right)$. The evaluation metrics are summarized in Table \ref{tab:metric} for details.

\section{Ablation on Model Architecture}
We evaluated varying hidden variables $d$ of text-VAE using the NYU Depth V2 dataset\cite{NYU-Depth-V2}, shown in Table \ref{tab:hidden_dim_supp}. A key consideration was ensuring the hidden space was sufficiently large to encode the necessary structural and geometric features for reconstructing depth maps. This size requirement arises from the need to preserve essential features about the scene's objects and layout derived from text features encoded by text-VAE.

However, it's equally crucial to avoid excessively large hidden variables. A relatively constrained dimensionality acts as a form of regularization, compelling the text-VAE to focus on extracting features crucial for depth decoding. Additionally, a limited hidden dimension prompts the model to learn not just the distribution mean but also its variance. This aspect is particularly important when mapping a text description to multiple scenes, such scenes' text features are encoded with identical distribution means but exhibit significant variance.

We established hidden variables $d$ of 32, 64, 128, 256, 512, and 1024 for training WorDepth. It was observed that the optimal hidden dimension is 128, striking a balance between capturing sufficient geometric features of scenes while maintaining effective regularization. Deviating from this optimal size, either too small or too large, adversely impacts performance.

\begin{table}[h]
\centering

\begin{adjustbox}{width=1.02\linewidth}
\hspace*{-2mm}
\begin{tabular}{l|c}
\toprule
\textbf{Metric}    & \textbf{Formulation} \\ 

\midrule
Abs Rel & $\frac{1}{N_e}\sum_{(i,j)\in \Omega}  \frac{|y^*(i,j) - {y}(i,j) |}{y^*(i,j)}$   \\
\midrule
RMSE    & $\sqrt{\frac{1}{N_e}\sum_{(i,j)\in \Omega} (y^*(i,j) - {y}(i,j))^2}$            \\
\midrule
$\log_{10}$   &    $\frac{1}{N_e}\sum_{(i,j) \in \Omega}  |\log_{10}(y^*(i,j)) - \log_{10}({y}(i,j)) | $ \\
\midrule
$\text{RMSE}_{\log}$   & $\sqrt{\frac{1}{N_e}\sum_{(i,j)\in \Omega} (\ln(y^*(i,j)) - \ln({y}(i,j)))^2} $ \\
\midrule
$\delta$          &    $\% \text { of }  y(i,j) \text { s.t. } \max (\frac{{y}(i,j)
}{y^*(i,j)}, \frac{y^*(i,j)}{{y}(i,j)}) <t h r \in [1.25,1.25^{2}, 1.25^{3}]$ \\
		
\bottomrule
\end{tabular}
\end{adjustbox}
\caption{\textbf{Evaluation metric for monocular depth estimation.} $y$ denotes predictions and $y^*$ denotes ground truth.}
\label{tab:metric}
\end{table}

\begin{table}[t!]
\centering
\LARGE
\begin{adjustbox}{width=\linewidth}
	\begin{tabular}{l|ccc|ccc}
	\toprule
		\textbf{Method}    & $\delta<1.25 \uparrow$ & $\delta<1.25^{2} \uparrow$ & $\delta<1.25^{3} \uparrow$ & Abs Rel $\downarrow$ & $\log _{10} \downarrow$ & RMSE $\downarrow$ \\ 

        \midrule 
        $d=32$    &0.925  &0.990  &\textbf{0.998}  &0.093  &0.039  &0.327      \\
        $d=64$    &0.928  &0.990  &\textbf{0.998}  &0.090  &0.039  &0.325       \\
        $\bm{d=128}$    & \textbf{0.932}  & \textbf{0.992} & \textbf{0.998} & \textbf{0.088} & \textbf{0.038} & \textbf{0.317}  \\
        $d=256$    & 0.930  & 0.991 & \textbf{0.998} &0.089 & 0.039 & 0.323 \\
        $d=512$    &0.929  &0.990  &\textbf{0.998}  &0.089  &0.039  &0.324        \\
        $d=1024$    &0.926 &0.989  &\textbf{0.998}  &0.091  &0.039  &0.325        \\
	\bottomrule
	\specialrule{0em}{7pt}{7pt}
	\end{tabular}
\end{adjustbox}
\caption{\textbf{Sensitivity to different numbers of hidden variables $d$.} Experiments are conducted on NYU Depth V2. $d$ is the number of hidden variables $d$ of the text-VAE.}
\label{tab:hidden_dim_supp}
\vspace*{-0.5cm}
\end{table}

\section{Additional Visualization on NYU Depth V2}
\label{sec:vis_nyu_supp}

\begin{figure*}[t!]
  \centering
    \includegraphics[width=1\textwidth]{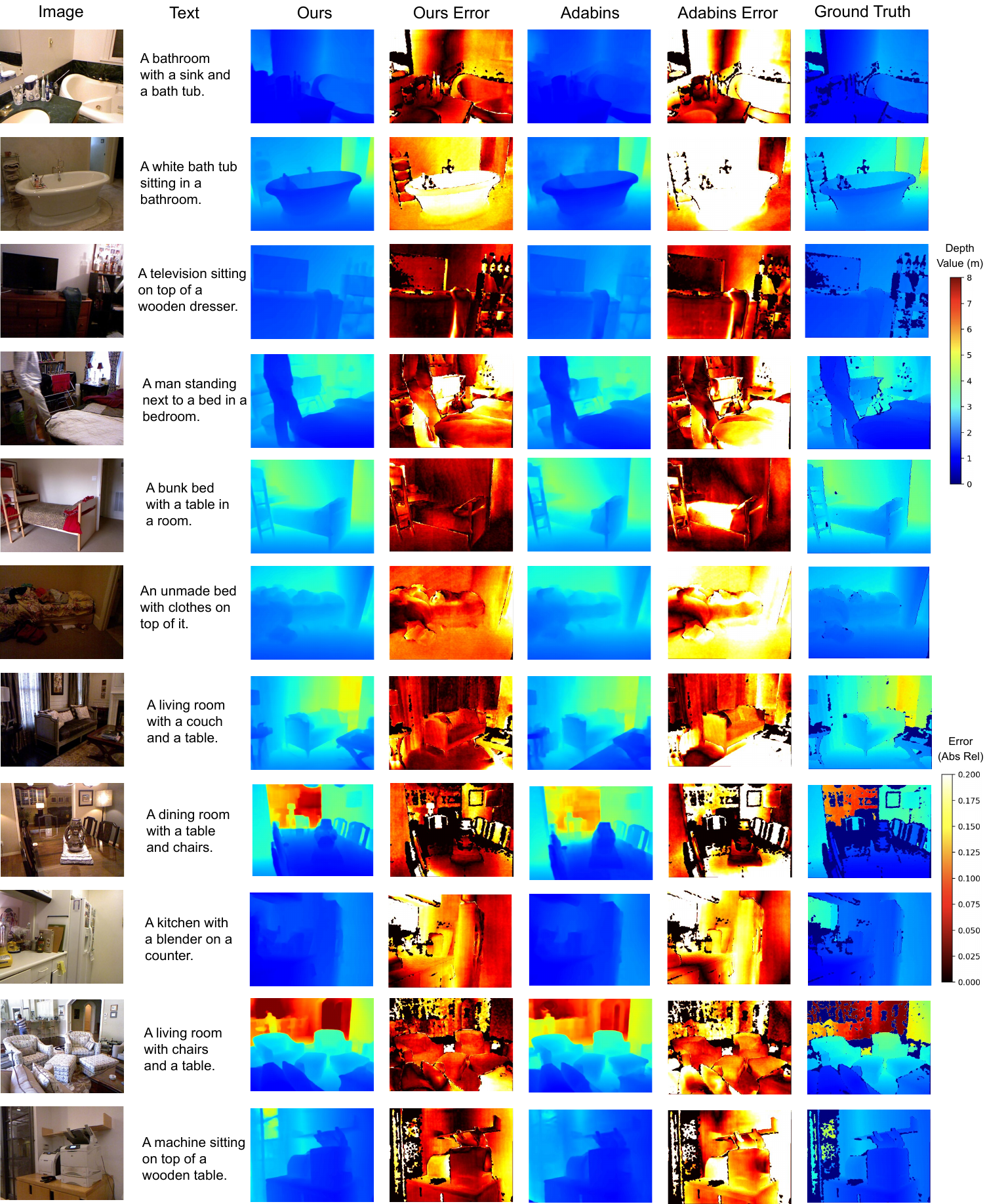}   \caption{\textbf{Additional visualization of monocular depth estimation on NYU Depth V2.}}
    \label{fig:vis_nyu_supp}
\end{figure*}

In this section, as illustrated in Figure \ref{fig:vis_nyu_supp}, We present additional visualizations comparing WorDepth with a baseline method AdaBins \cite{bhat2021adabins} on the NYU Depth V2~\cite{NYU-Depth-V2} dataset, emphasizing the advantages gained from integrating the language prior. Compared with AdaBins, the error map, with its brighter regions highlighting larger errors, clearly demonstrates that WorDepth achieves more precise depth predictions for objects identified in the text description. For instance: ``a sink and a bath tub'' in the first row, ``a white bath tub'' in the second row, ``a wooden dresser'' in the third row, ``a bed'' in the fourth row, ``a bunk bed'' in the fifth row,  ``an unmade bed with clothes on top of it" in the sixth row, ``a couch and a table" in the seventh row, ``a table and chairs" in the eighth row, ``a blender on a counter" in the ninth row, ``chairs" in the tenth row, and ``machine on top of a wooden table" in the last row.


\vspace{0.5cm}


\vspace{0.5cm}
\begin{figure*}[t!]
  \centering
    \includegraphics[width=\textwidth]{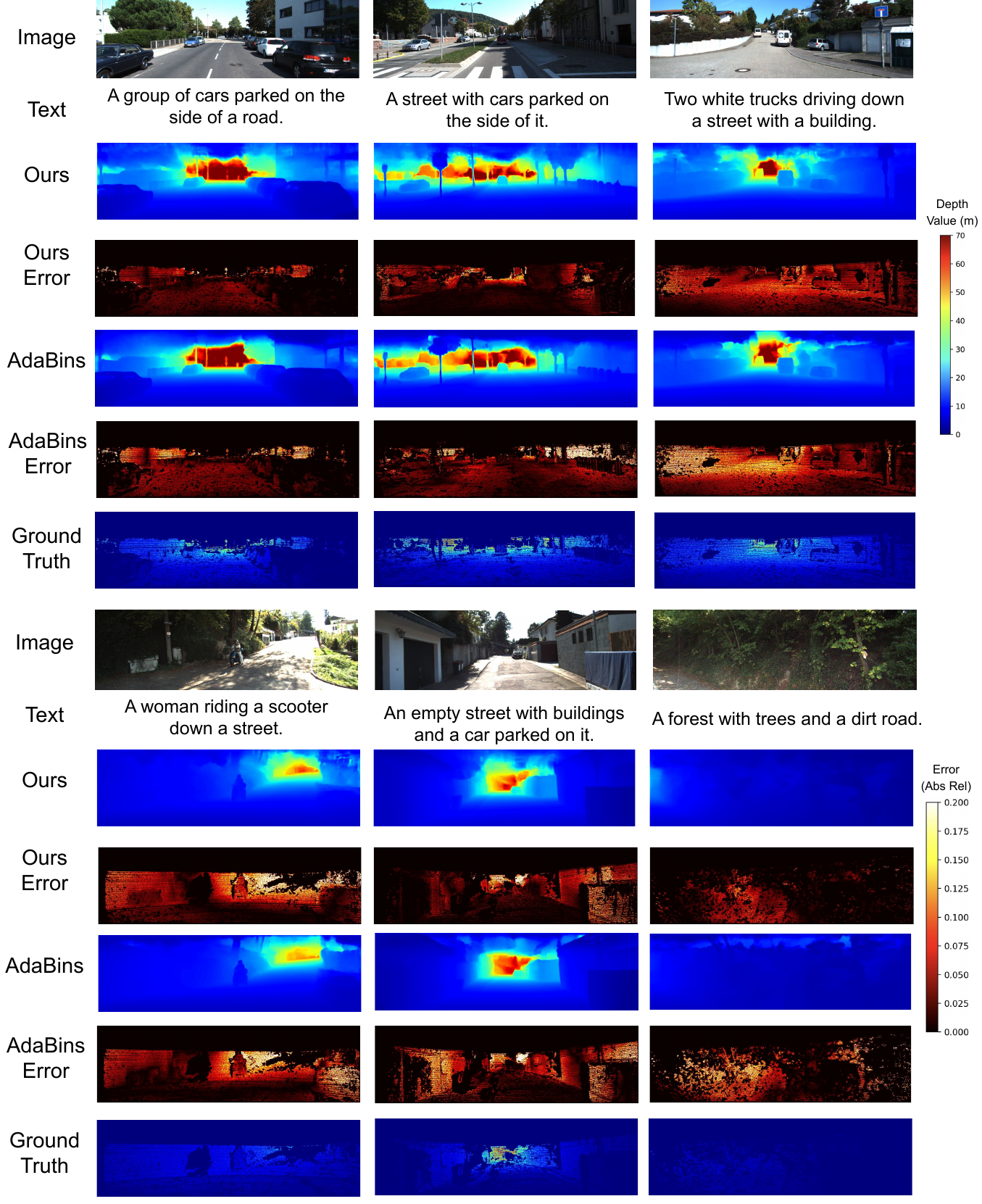}   \caption{\textbf{Additional visualization of monocular depth estimation on KITTI Eigen split. }}
    \label{fig:vis_kitti_supp}
\end{figure*}

\section{Additional Visualization on KITTI}
\label{sec:vis_kitti_supp}
This section, depicted in Figure \ref{fig:vis_kitti_supp}, showcases visualizations of Monocular Depth Estimation in outdoor scenarios with the KITTI dataset \cite{KITTI} using Eigen Split \cite{Eigen-Split}, comparing with Adabins \cite{bhat2021adabins}. 
Due to the limited variety of objects in outdoor scenes, our method captures fewer objects compared to indoor scenes. However, when salient objects and scenes are present outdoors, our method gains a preliminary understanding of their scale. This understanding aids in enhancing monocular depth estimation for these objects. The error map's brighter regions, which emphasize greater absolute relative errors, unequivocally show that WorDepth outperforms AdaBins in making more accurate depth predictions for objects and scenes mentioned in the text description. For instance: ``two white trucks" in the upper right, ``a woman riding a scooter" in the lower left, ``buildings" in the lower middle, and ``forest with tree" in the lower left.

\end{document}